\documentclass{article}

\usepackage{arxiv}

\usepackage[utf8]{inputenc} 
\usepackage[T1]{fontenc}    
\usepackage{hyperref}       
\usepackage{url}            
\usepackage{booktabs}       
\usepackage{amsfonts}       
\usepackage{nicefrac}       
\usepackage{microtype}      
\usepackage{lipsum}
\usepackage{amsmath}
\usepackage{fixltx2e}
\usepackage{hyperref}
\usepackage{graphicx}

\title{Modeling severe traffic accidents with spatial and temporal features}

\author{
  Devashish Khulbe \\
  New York University\\
  New York, NY 10003 \\
  \texttt{dk3596@nyu.edu} \\
   \And
  Soumya Sourav \\
  University of Texas at Dallas\\
  Dallas, TX 75080 \\
  \texttt{sxs180011@utdallas.edu} \\
}

\begin{document}
\maketitle

\begin{abstract}
We present an approach to estimate the severity of traffic related accidents in \textit{aggregated} (area-level) and \textit{disaggregated} (point level) data. Exploring spatial features, we measure 'complexity' of road networks using several area level variables. Also using temporal and other situational features from open data for New York City, we use Gradient Boosting models for inference and measuring feature importance along with Gaussian Processes to model spatial dependencies in the data. The results show significant importance of 'complexity' in aggregated model as well as as other features in prediction which may be helpful in framing policies and targeting interventions for preventing severe traffic related accidents and injuries.
\end{abstract}


\section{Introduction}
Traffic related accidents contribute to the deaths of around 1.35 million people and injuries to around 30 million people worldwide. 93\% of the world's fatalities on the roads occur in low- and middle-income countries, even though these countries have approximately 60\% of the world's vehicles \cite{who}. The above fact is informative regarding the possible link between street conditions and design and fatal accidents. Additionally, road traffic crashes cost most countries 3\% of their gross domestic product, indicating that curbing traffic accidents is financially important. With around 60\% of the world population predicted to live in cities by 2030, making urban areas safe for pedestrians and vehicles simultaneously is an important area to delve into. Further, assessing the role of multiple location and time factors in prediction can help the concerned authorities in deploying targeted interventions for public safety. With real-time large open data of accidents and information about urban network of streets, we argue that severity of accidents can be estimated using modern Machine Learning (ML) techniques. Recently, ML has proved to be an important tool for predicting traffic accidents and crash severity, with a variety of tools being used for accident risk prediction \cite{intro2}. In this paper, we present approaches to infer injury-related and fatal accidents for area level and observation level data for New York City. We also introduce a new feature measuring the complexity of area-level street networks. We model the spatial autocorrelation in the data in the area level model which the regular model may not be able to learn. Interpretable techniques like Gradient Boosting are used to measure feature importance of our variables, the results of which show impact of several spatial and temporal features in inference. \\

This work contributes to the current research by introducing some new significant predictors and presenting a way to account for spatial dependencies in accidents data.

\subsection{Related Work}
Significant amount of literature can be attributed to modeling traffic accident and their severity using diverse set of variables. Features like curvature, road width, urban/rural area and gender of driver area have been shown to be significant in accident modeling \cite{lit1}. Another work \cite{lit2} describe models for predicting the expected number of accidents at urban junctions and road links as accurately as possible, explaining 60\% of variation for road links. This is indicative of importance of location based features of streets and junctions in accident modeling. An approach \cite{lit3} models severe accidents for area level predictions using linear model using features like intersection density, vehicle speed and number of households which turned out to be significant. This work also uses Geographic Weighted Regression (GWR) to account for spatial correlation. Deep learning approaches have also been proposed for modeling traffic accidents \cite{lit7} in the past but we aim to build interpretable models in order to measure importance of features with this paper. Our work aims to further introduce and use new predictors to model and subsequently use non-linear models to predict severe accidents both for area-level and point-level data and also account for possible spatial dependencies for area-level data.

\section{Data \& Methods}
\label{methods}

\paragraph{Traffic Accidents}
We use open data for New York City maintained by New York City Police Department (NYPD). The data contains entries of motor vehicle related accidents, containing their coordinates, date and time of incident, type of vehicles involved and number of injuries and deaths. For the aggregated model, we aggregate the number of accidents on census tract level for the city. The model is thus a regression problem with the number of severe incidents as target. For the disaggregated model, the problem is essentially of binary classification type where we aim to classify a traffic accident as severe or non-severe. We define the severity as any incident where number of injuries or deaths are equal to greater than 1. We assign binary values as the target:

\[ \begin{cases} 
      0, & injuries/deaths = 0 \\
      1, & otherwise \\
   \end{cases}
\]

Considering the data from July 2011 till May 2019, this results in a total of around 1,000,000 non-severe accidents and around 200,000 severe accidents, indicating a problem of imbalance which we address further in this section.

\paragraph{Spatial autocorrelation}
Assuming that our data is not \textit{i.i.d}, we measure local spatial autocorrelation for the number of severe accidents y\textsubscript{j} for census-tract aggregated level through the \textit{Local Moran's I} statistic calculated as:

\begin{equation}
I_j = (n-1)\frac{y_j - \overline{y}}{\sum _{k=1, k\neq j}^{m} w_{j,k}(y_k - \overline{y})^2}\sum _{k=1, k\neq j}^{m}w_{j,k}(y_k - \overline{y})
\end{equation}

where where n is the number of spatial units indexed by i and j; $\overline{y}$ represents the mean of y\textsubscript{j} and w represents the spatial weight between the features j and
k. The weights for neighbouring and non-neighbouring areas of each census tract j are taken as:  w\textsubscript{j, k} = 1 if k is a queen contiguous neighbour of j and w\textsubscript{j, k} = 0 otherwise.

\begin{figure}[h]
\caption{Local spatial autocorrelation of total severe accidents in New York (p<0.05)}
\centering
\includegraphics[width=7cm, height=6cm]{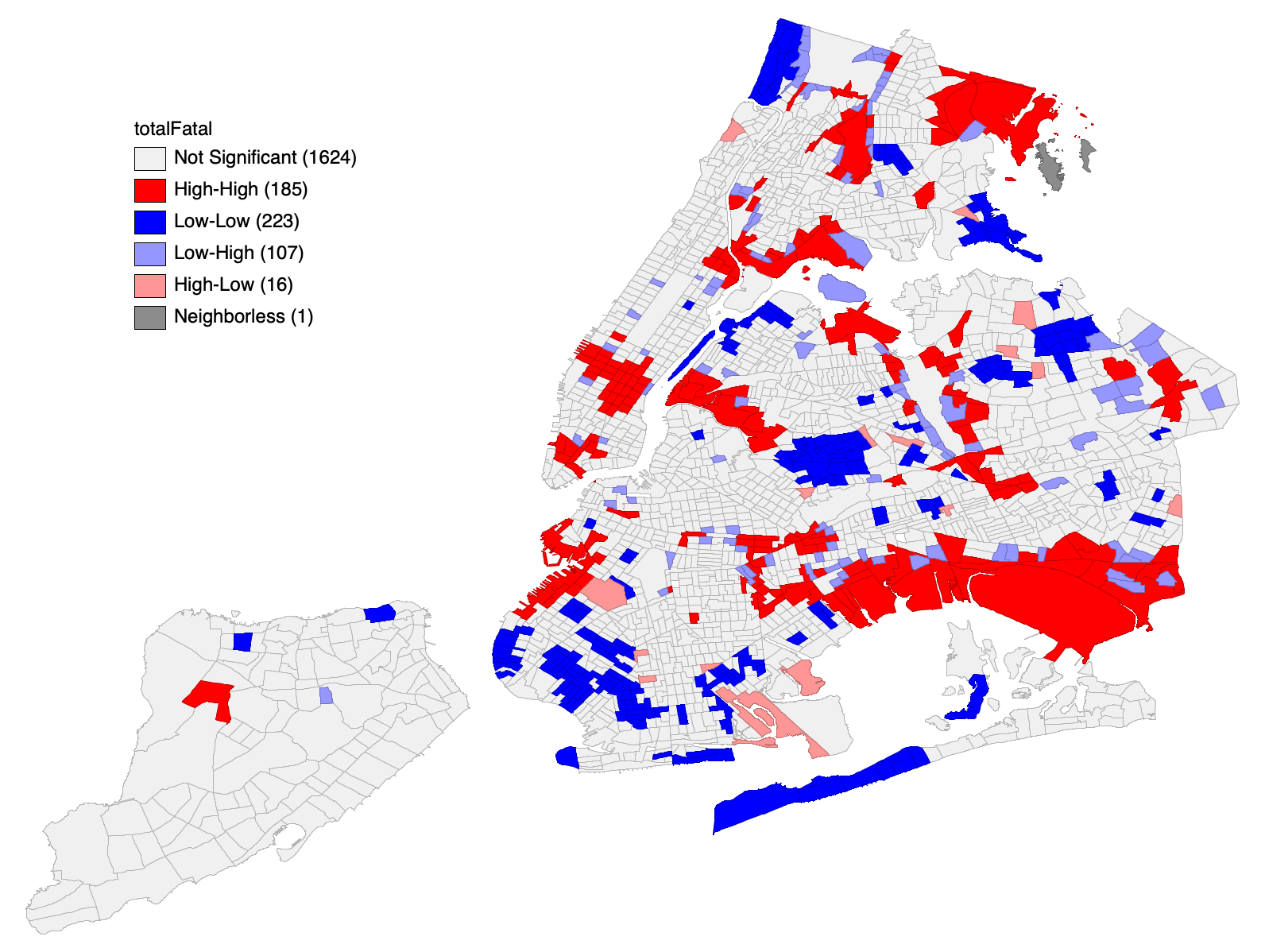}
\label{autocorr}
\end{figure}

\paragraph{Complexity}
Introducing a new feature as a proxy for 'complexity' of street level networks, we define it as a multiple of number of intersections and circuity of a given network. We then measure complexity for our all networks of census tract areas. The circuity is defined as a ratio of network distance to Euclidean distance for a given street network. Considering that number of intersections has been show to be correlated to accidents in previous work \cite{lit3}, multiplying it with circuity account for further intricacies of the network. Thus, our new feature measures complexity of the networks by accounting for factors like number of turns, intersections and nodes.

\paragraph{Other Variables}
Other features we consider are average street width, vehicle types, average number of bike lanes, day of week and time of day of incident, for which only the hour value is taken. Subsequently, the census-tract level aggregated data set results in 2156 observations and the point level (disaggregated) data set contains around 1,200,000 observations.

\begin{table}[h!]
  \begin{center}
    \caption{Aggregated (census-tract level) data set}
    \label{tab:table1}
    \begin{tabular}{l|c|r} 
      \textbf{Features (observations)} & \textbf{Mean value} & \textbf{Std. Deviation}\\
      \hline
      Complexity (2156) & 30.58 & 39.34\\
      Avg. Street width in meters (2156) & 34.13 & 5.85\\
      Avg. Bike lanes (2156) & 1.47 & 1.35\\
      Avg. node degree (2156) & 3.59 & 0.83\\
    \end{tabular}
  \end{center}
\end{table}

\paragraph{Data imbalance}
Imbalanced data set is the one which suffers from the problem of classes not being in proportion. This causes a machine learning model to generate fake accuracy reports with the imbalanced data set. With our model we have tried to evade this problem by the use of SMOTE, since our data is fairly imbalanced with positive class accounting for just about 23\% of the total observations.

SMOTE (Synthetic Minority Over-Sampling Technique) - it is an oversampling method which can create synthetic samples from minor classes instead of just copying them. The algorithm selects two or more similar instances (using a distance measure) and changing an observation one feature at a time by a random amount within the difference to the neighboring data points.

\section{Results}

\begin{figure}[h]
\centering

  \centering
  \caption{Hour of day (x-axis) and number of severe accidents (y-axis)}
  \includegraphics[width=.42\linewidth]{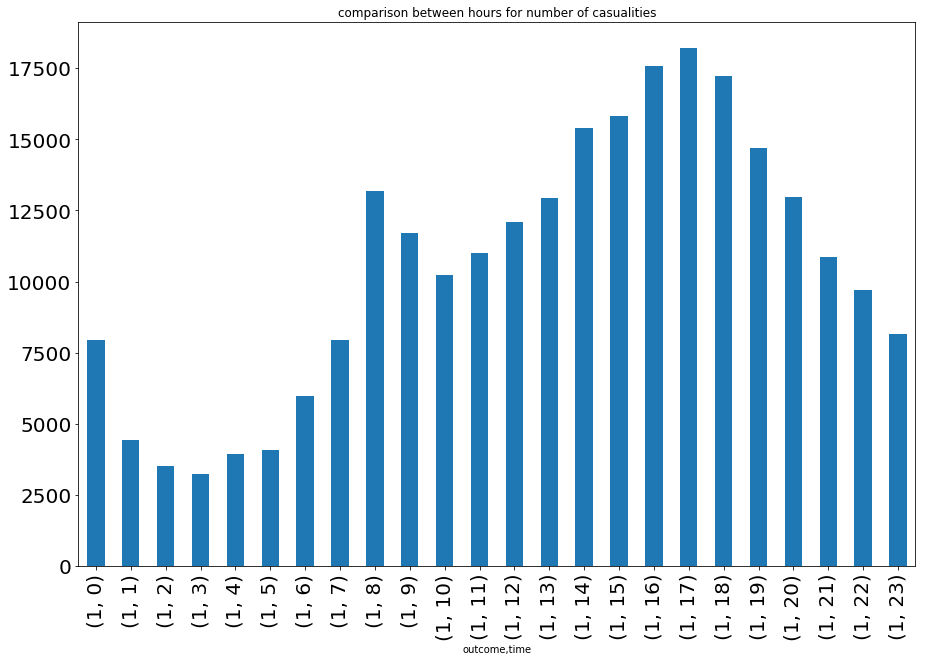}
  
  \label{fig:test1}
  
\end{figure}
\begin{figure}[h]
    \centering
    \caption{Day of week (x-axis) and number of severe accidents (y-axis)}
    \includegraphics[width=.42\linewidth]{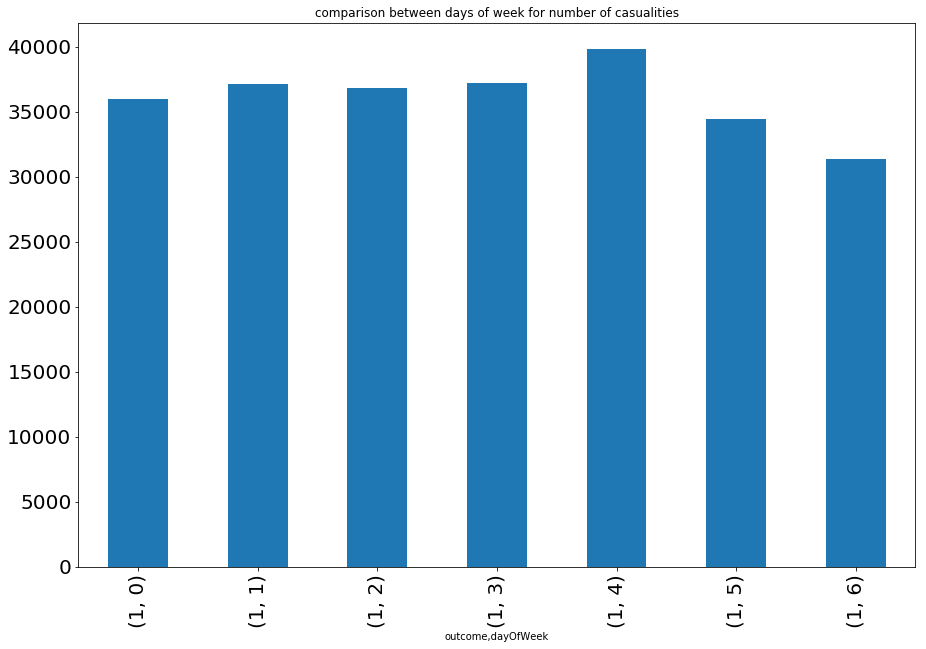}
    \label{fig:my_label}
\end{figure}

\subsection{Spatial \& Temporal Features}

We observe that hour of day of the accident may be important as a predictor, with majority of incidents happening in the evening and night hours. Also, we consider day of week as a predictor, where we observe that weekdays have a slightly greater proportion of accidents as compared to weekends. 
Also, we decide to model the spatial autocorrelation observed in the data for the \textit{aggregated model} since on observation, there seem to be spatial dependencies in the accidents with some areas having high proportion of incidents.

\subsection{Aggregated model}
For the census-tract level model, we estimate the number of severe accidents as a function of spatial features \textit{s} described in table \ref{tab:table1} as \textit{y = f(s) + $\epsilon$}, where $\epsilon$ is the error term. Further, we model the error $\epsilon$ using spatial context \textit{$\epsilon$ = g(x)}, where \textit{x} are the centroids of the spatial area we are considering. We use Gradient Boosting (GB) model using 20 fold cross validation for the first part and Gaussian Processes with Radial Basis Function (RBF) kernel for modeling the residual error term. Results show that the first step explains around 34\% variation (measured by the $R^2$ value) in the data and further 13\% is explained by modeling the residual error term. Looking at the feature importance, complexity turns out to be the most important in prediction followed by average number of nodes, average street width and average number of bike lanes in the census tract.

\begin{table}[h!]
  \begin{center}
    \caption{Aggregated model}
    \label{tab:table2}
    \begin{tabular}{l|c|r} 
      \textbf{Step} & \textbf{Model used} & \textbf{$R^2$}\\
      \hline
      \textit{y = f(s) + $\epsilon$}  & Gradient Boosting & 0.338\\
      \textit{$\epsilon$ = g(x)} & Gaussian Process & 0.132\\
    \end{tabular}
  \end{center}
\end{table}

\subsection{Disaggregated model}

\begin{figure}[h]
    \centering
    \caption{Receiver Operator Characteristic (ROC) curve for classification models}
    \includegraphics[scale=0.25]{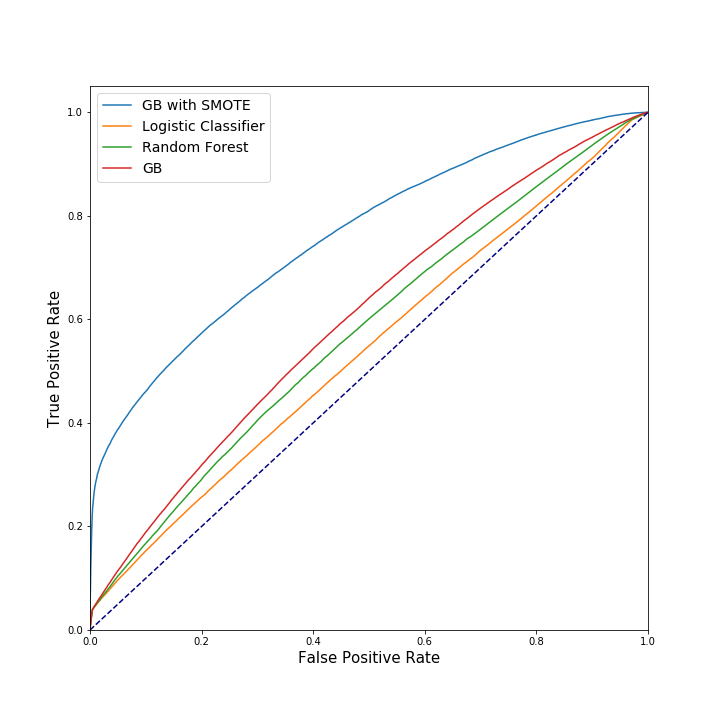}
    \label{fig:my_label}
\end{figure}

For the \textit{disaggregated} data, the goal to classify each data point as severe or not severe based on the features. The data is thus trained as a binary classification problem with four classifiers with 10-fold cross validation. Oversampling the positive class (severe) points with SMOTE and then training the resulting data with Gradient Boosting results in highest AUC score of 0.72. Along with the temporal features like time, the spatial features are also used which are attributed to the census-tract in which the incident happened.

\begin{table}[h!]
  \begin{center}
    \caption{Disaggregated (point classification) model}
    \label{tab:table3}
    \begin{tabular}{l|c|r} 
      \textbf{Model} & \textbf{AUC}\\
      \hline
      Gradient Boosting with SMOTE & 0.729\\
      Gradient Boosting & 0.604\\
      Random Forests & 0.575\\
      Logistic Regression & 0.539\\
    \end{tabular}
  \end{center}
\end{table}

The important features for classification include class of vehicle (specifically whether it was a two-wheeler or truck), complexity of the area and the hour of the incident.

\section{Discussion}
We observe that spatial features which account for complexity like number of intersections, network nodes, circuity along with vehicle type are important features in prediction of severe accidents. Complexity turns out to be the most important predictor in the \textit{aggregated} model while it it fairly good in the point level classification too. It is interesting to note that average number of bike lanes is not one of the most important predictors for injury and fatal accident classification, despite one of the research concluding that bike lanes make a route safer \cite{lit4}. This may be because we took the average number of bike lanes in a neighborhood while making predictions. Maybe a more fine street level feature information on this can change our results, which may be a scope for future work in this work. We also notice that temporal features like day of week and hour of day when incident happened are important information about accidents in general. Most of the accidents take place over the weekdays and during the evening and late-night hours. Though, these temporal features do not contribute much in our predictive model, they are good predictors of vehicle collisions and accidents. The spatial features inform us from a road and street design perspective in a neighborhood while the temporal variables along with information about vehicle in the accidents can be important from a real-time emergency deployment viewpoint.

\section{Conclusion}
In this work, we presented an approach using Machine Learning techniques to model non-fatal (injury) and fatal (death) traffic accidents in urban environments using spatial and temporal variables. We found the importance of factors like street width, vehicle type, time of day and the new created feature 'complexity' of a street network in prediction of severe accidents both in the area level and point level data. This information can be critical in implementation of policies regarding construction and design of neighborhood streets and can help in effective operation of emergency services.

Future work in this domain can be extended to incorporate other socio-economic features in prediction and determination of most affected demographics in terms of road traffic incidents.

\bibliographystyle{unsrt}  


\begin{thebibliography}{1}

\bibitem{who}
Road traffic injuries.
\newblock https://www.who.int/news-room/fact-sheets/detail/road-traffic-injuries.

\bibitem{intro2}
Amirfarrokh Iranitalab, Aemal Khattak.
\newblock Comparison of four statistical and machine learning methods for crash severity prediction, Accident Analysis \& Prevention.
\newblock Volume 108, 2017, Pages 27-36, ISSN 0001-4575, https://doi.org/10.1016/j.aap.2017.08.008.

\bibitem{lit1}
Mohamed A. Abdel-Aty, A.Essam Radwan.
\newblock Modeling traffic accident occurrence and involvement,
Accident Analysis \& Prevention.
\newblock  Volume 32, Issue 5, 2000, Pages 633-642, ISSN 0001-4575, https://doi.org/10.1016/S0001-4575(99)00094-9.

\bibitem{lit2}
Poul Greibe.
\newblock Accident prediction models for urban roads,
Accident Analysis \& Prevention,
\newblock  Volume 35, Issue 2, 2003, Pages 273-285, ISSN 0001-4575,
https://doi.org/10.1016/S0001-4575(02)00005-2.

\bibitem{lit3}
Hadayeghi, A., Shalaby, A. S., \& Persaud, B. (2003). 
\newblock Macrolevel Accident Prediction Models for Evaluating Safety of Urban Transportation Systems. 
\newblock Transportation Research Record, 1840(1), 87–95. https://doi.org/10.3141/1840-10

\bibitem{lit4}
Harris MA, Reynolds CCO, Winters M, et al.
\newblock Comparing the effects of infrastructure on bicycling injury at intersections and non-intersections using a case–crossover design.
\newblock Injury Prevention 2013;19:303-310.

\bibitem{lit5}
Persaud, B., Lord, D., \& Palmisano, J. (2002). 
\newblock Calibration and Transferability of Accident Prediction Models for Urban Intersections. 
\newblock Transportation Research Record, 1784(1), 57–64. https://doi.org/10.3141/1784-08

\bibitem{lit6}
El-Basyouny, K., \& Sayed, T. (2006).
\newblock Comparison of Two Negative Binomial Regression Techniques in Developing Accident Prediction Models. 
\newblock Transportation Research Record, 1950(1), 9–16. https://doi.org/10.1177/0361198106195000102

\bibitem{lit7}
Abdelwahab, H. T., \& Abdel-Aty, M. A. (2001). 
\newblock Development of Artificial Neural Network Models to Predict Driver Injury Severity in Traffic Accidents at Signalized Intersections.
\newblock Transportation Research Record, 1746(1), 6–13. https://doi.org/10.3141/1746-02

\bibitem{lit8}
Mohammed A. Quddus.
\newblock Time series count data models: An empirical application to traffic accidents, Accident Analysis \& Prevention.
\newblock Volume 40, Issue 5, 2008, Pages 1732-1741, ISSN 0001-4575,
https://doi.org/10.1016/j.aap.2008.06.011.

\bibitem{lit9}
Z Sawalha and , T Sayed.
\newblock Traffic accident modeling: some statistical issues
\newblock Canadian Journal of Civil Engineering, 2006, 33(9): 1115-1124, https://doi.org/10.1139/l06-056

\end{thebibliography}

\end{document}